\documentclass[letterpaper]{sig-alternate-2013}

\usepackage[
  pass,
]{geometry}

\permission{\copyright 2017 International World Wide Web Conference Committee (IW3C2), 
published under Creative Commons CC BY 4.0 License.}
\conferenceinfo{WWW 2017,}{April 3--7, 2017, Perth, Australia.}
\copyrightetc{ACM \the\acmcopyr}
\crdata{978-1-4503-4913-0/17/04. \\
http://dx.doi.org/10.1145/3038912.3052611 \\
\includegraphics{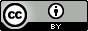}
}
 
\usepackage{times}
\usepackage{latexsym}
\usepackage{capt-of}

\usepackage{times}
\usepackage{latexsym}
\usepackage{graphicx}
\usepackage{mdwlist}
\usepackage{amsmath}
\usepackage{amssymb}
\usepackage{color}
\usepackage{caption}
\usepackage{subcaption}
\usepackage{url}
\usepackage{booktabs}
\usepackage{tabularx}
\usepackage{array}
\usepackage{multirow}

\usepackage{hyperref}

\newcommand{\R}{{\mathbb{R}}}
\newcommand{\comment}[1]{}

\renewcommand{\captionfont}[1]{\footnotesize{#1}}
\newcommand{\ntweets}{$300M$~}

\begin{document}

\title{Leveraging Large Amounts of Weakly Supervised Data for Multi-Language Sentiment Classification}

%
%
%
%
%

\numberofauthors{8} 
%
\author{
%
%
\alignauthor
Jan Deriu\\
       \affaddr{ZHAW}
\alignauthor
Aurelien Lucchi \\
	\affaddr{ETH Zurich}
\alignauthor
Valeria De Luca \\
	\affaddr{ETH Zurich}
\and   
\alignauthor
Aliaksei Severyn \\
	\affaddr{Google Research}
\alignauthor
Simon M\"uller \\
	\affaddr{SpinningBytes AG}
\alignauthor
Mark Cieliebak \\
	\affaddr{SpinningBytes AG}
\and   
\alignauthor Thomas Hofmann \\
	\affaddr{ETH Zurich}
\alignauthor Martin Jaggi \\
	\affaddr{EPFL}
}

\date{3 April 2017}

\maketitle
\begin{abstract}
This paper presents a novel approach for multi-lingual sentiment classification in short texts. This is a challenging task as the amount of training data in languages other than English is very limited. Previously proposed multi-lingual approaches typically require to establish a correspondence to English for which powerful classifiers are already available. In contrast, our method does not require such supervision. 
We leverage large amounts of weakly-supervised data in various languages to train a multi-layer convolutional network and demonstrate the importance of using pre-training of such networks. We thoroughly evaluate our approach on various multi-lingual datasets, including the recent SemEval-2016 sentiment prediction benchmark (Task~4), where we achieved state-of-the-art performance. 
We also compare the performance of our model trained individually for each language to a variant trained for all languages at once. We show that the latter model reaches slightly worse - but still acceptable - performance when compared to the single language model, while benefiting from better generalization properties across languages.
\end{abstract}



\section{Introduction}

Automatic sentiment analysis is a fundamental problem in natural language processing (NLP). A huge volume of opinionated text is currently available on social media. On Twitter alone, 500 million tweets are published every day. Being able to manually process such a high volume of data is beyond our abilities, thus clearly highlighting the need for automatically understanding the polarity and meaning of these texts. 
Although there have been several progresses towards this goal, automatic sentiment analysis is still a challenging task due to the complexity of human language, where the use of rhetorical constructions such as sarcasm and irony easily confuse sentiment classifiers. Contextualization and informal language, which are often adopted on social media, are additional complicating factors. The Internet is also multi-lingual and each language has its own grammar and syntactic rules.

Given all these difficulties, it is not surprising that the performance of existing commercial systems is still rather poor, as shown in several recent studies~\cite{cieliebak2013potential,ribeiro2015benchmark}. The benchmark work of Ribeiro \textit{et al.}~\cite{ribeiro2015benchmark} showed that even the performance of the best systems largely varies across datasets and overall leaves much room for improvement. Hence it is important to design a method that generalizes well to different domains and languages.
\\

\textbf{Contributions.}
The majority of current research efforts in sentiment analysis focuses on the English language. This is partially due to the large number of resources available in English, including sentiment dictionaries, annotated corpora and even benchmark datasets. An example is the SemEval competition, which is one of the largest competitions on semantic text evaluation and covers several tasks for sentiment analysis \cite{SemEval:2016:task4}.

However, only 26.3\% of the total number of internet users in 2016 are English speakers ~\cite{website:internetstats} and only 34\% of all tweets are written in English \cite{website:twitterlanguages}. Hence there is a strong incentive to develop methods that work well with other languages. In this work, we focus on the question of how sentiment analysis can be done for multiple languages by leveraging existing technologies. Our method is the state-of-the-art approach for sentiment analysis on Twitter data which recently won the SemEval-2016 competition~\cite{deriu2016swisscheese}. Here we additionally explore how to best adapt this approach to other languages. The core component of our system is a multi-layer convolutional neural network (CNN), trained in three phases: i) unsupervised phase, where word embeddings are created on a large corpus of unlabeled tweets; ii) distant supervised phase, where the network is trained on a weakly-labeled dataset of tweets containing emoticons; and iii) supervised phase, where the network is finally trained on manually annotated tweets. For English, this system achieved an F1-score of 62.7\% on the test data of SemEval-2016~\cite{deriu2016swisscheese}. 

Although existing CNN approaches
~\cite{Severyn:2015kt, deriu2016swisscheese} can a-priori be trained on any language other than English, these nevertheless require a large amount of training data. Yet resources in languages other than English are lacking, and manually labeling tweets is a time-consuming and expensive process. Two straightforward solutions that do not require manual work can be envisioned: (1) automatically translate the data into English and run the existing English classifier; or (2) train a CNN using only weakly-labeled tweets without using any supervised data. It is expected that a fully-trained CNN would perform better than the aforementioned cases. However, it is unclear if such improvement is significant and justifies the need of manually labeling thousands of tweets.

In this paper, we investigate how to effectively train and optimize a CNN for multi-lingual sentiment analysis. We compare the performance of various approaches for non-English texts. In details, our main contributions are:

\begin{itemize}
\itemsep0em
\item An evaluation of the state-of-the-art CNN approach similar to the one proposed in~\cite{deriu2016swisscheese} on three new languages, namely French, German and Italian
\item A thorough analysis of the influence of network parameters (number of layers, hyper-parameters) and other factors, e.g. the amount of distant-supervised and supervised data, on end-to-end performance
\item For each language, a comparison of various approaches for sentiment analysis: (i) full training of the CNN for the considered language; and (ii) automatically translating the texts into a language (English) where a sentiment classifier already exists. Other baseline methods, described in the experimental section, are also compared
\item In addition, we show that a single CNN model can be successfully trained for the joined task on all languages, as opposed to separate networks for each individual language. This approach has the advantages of removing the reliance on (possibly inaccurate) language identification systems and it can be easily extended to new languages and multi-language texts. We provide detailed comparison to similar per-language models, and show that the proposed joint model still performs relatively well
\item Public release of the source code as well as pre-trained models for all languages tested in this paper, on\\ \url{http://github.com/spinningbytes/deep-mlsa}
\\
\end{itemize}


\section{Related work}

In the following, we provide an overview of the most relevant works, related to 
the application of neural networks to sentiment classification, distant supervision and training multi-lingual text classifiers.
\\

\textbf{Neural networks.}
Neural networks have shown great promise in NLP over the past few years. Examples are in semantic analysis \cite{shen2014latent}, machine translation \cite{gao2014learning} and sentiment analysis \cite{socher2013recursive}.
In particular, shallow CNNs have recently improved the state-of-the-art in text polarity classification demonstrating a significant increase in terms of accuracy compared to previous state-of-the-art techniques~\cite{Kim:2014vt,Kalchbrenner:2014wl,dosSantos:2014tr,Severyn:2015tb,Johnson:2015wl,Rothe:2016te,deriu2016swisscheese}. These successful CNN models are characterized by a set of convolution filters acting as a sliding window over the input word sequence, typically followed by a pooling operation (such as max-pooling) to generate a fixed-vector representation of the input sentence. 
\\

\textbf{CNNs vs RNNs.} Recently, recurrent neural network architectures (RNNs), such as long short-term memory networks (LSTMs), have received significant attention for various NLP tasks. Yet these have so far not outperformed convolutional architectures on polarity prediction \cite[Table 4]{Semeniuta:2016ww}. This has been evidenced by the recent SemEval-2016 challenge \cite{SemEval:2016:task4}, where systems relying on convolutional networks rank at the top. In fact, long-term relationships captured well by LSTMs are of minor importance to the sentiment analysis of short tweets. On the contrary, learning powerful $n$-gram feature extractors (which convolutional networks handle very well) contributes much more to the discriminative power of the model, since these are able to effectively detect sentiment cues. Additionally, LSTMs are much more computationally expensive than CNNs, preventing their application to very large collections like the one used in this paper (hundreds of millions tweets).
\\

\textbf{Distant-supervised learning.}
The use of semi-supervised or unsupervised learning has been an active research direction in machine learning and particularly for various NLP applications. There is empirical evidence that unsupervised training can be beneficial for supervised machine learning tasks~\cite{erhan2010does}. 
In this paper, we consider a variant of unsupervised learning named distant pre-training which consists in inferring weak labels from data without manual labels. This approach has been used for text polarity classification where significantly larger training sets were generated from texts containing emoticons~\cite{Go:2009ut,Severyn:2015tb}. Severyn and Moschitti~\cite{Severyn:2015tb} have shown that training a CNN on these larger datasets, followed by additional supervised training on a smaller set of manually annotated labels, yields improved performance on tweets.
\\

\textbf{Multi-language sentiment classification.}
Sentiment classification has drawn a lot of attention in the past few years both in industry and academia~\cite{SemEval:2016:task4,cieliebak2013potential}. Yet most of the research effort has been focusing on tweets written in one language (mostly English). 
One exception is the work of Boiy and Moens ~\cite{boiy2009machine} that studied the portability of a learned sentiment classification model across domains and languages. They focused on French, Dutch and English, and showed that significant disparities between these languages can severely hinder the performance of a classifier trained on hand-crafted features.

The major factor that limits the development of accurate models for multi-lingual sentiment analysis is the lack of supervised corpora~\cite{balahur2012multilingual,dashtipour2016}.
Most of the existing approaches addressing this problem~\cite{mihalcea2007learning,araujo2016evaluation} try to transfer knowledge from English -- for which tools, labelled data and resources are abundant --  to other languages for which resources are rather limited. An example is the approach introduced in~\cite{mihalcea2007learning}, which transfers hand-crafted subjectivity annotation resources - such as a per-word sentiment lexicon - from English to Romanian. 
A similar approach introduced in~\cite{araujo2016evaluation} consists in translating a target language to English and then to use an English sentiment classifier rather than one specific to the target language. Several approaches have also been proposed to build distributed representations of words in multiple languages. The work of Wick \textit{et al.}~\cite{wick2015minimally} used a Wikipedia corpus of five languages to train word embeddings, and then used anchor terms (names, cross-lingual words) to align the embeddings. Gouws \textit{et al.}~\cite{gouws2015bilbowa} proposed a method to create bilingual word vectors by requiring words that are related across the two languages.

All the aforementioned approaches 
rely on having access to a set of correspondences between English and the target language. Some of these methods also require translating the target language to English. Yet machine translation is a very challenging task in NLP and represents an additional source of error in the classification system, due to various problems such as sparseness and noise in the data~\cite{dashtipour2016}. Furthermore, such methods crucially rely on accurate language identification, which is a very difficult task, especially on short texts. See e.g.~\cite{lui2012langid,lui2011cross} for an overview of these methods and their limitations in generalizing to different domains.

In this work, we also investigate the performance of a language-independent classifier consisting of a CNN trained on all languages at once. This approach is similar to the Na\"ive Bayes classifier proposed in~\cite{narr2012language}, excepts that it relies on simple hand-crafted word-level features instead of the CNN architecture used in this work.

\begin{figure*}[t]
  \centering
  \includegraphics[width=0.95\textwidth]{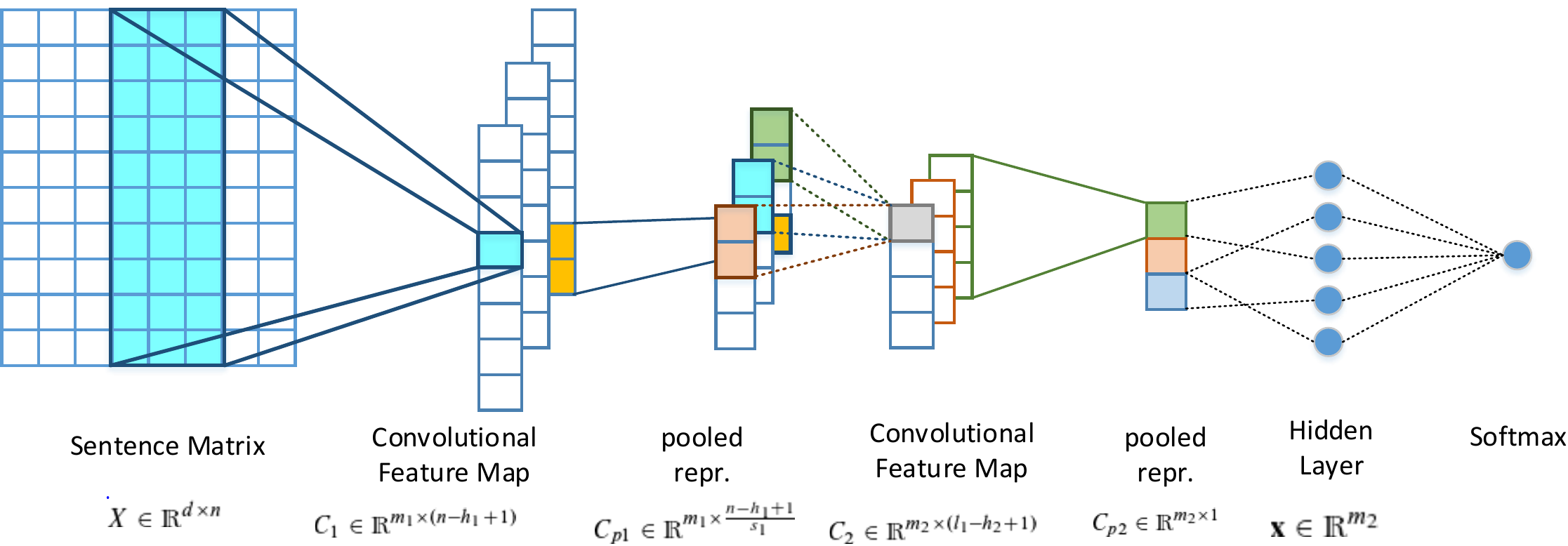} 
  \caption{Architecture of the proposed CNN model with 2 convolutional layers}
  \label{fig:cnn}
\end{figure*}

\section{Model}
\label{sec:model}

Our model follows a multi-layer CNN architecture, which we firstly introduced in \cite{deriu2016swisscheese}. Given an input sequence of words, the corresponding sequence of word embeddings is fed as input to the first 1d convolutional layer. Each convolutional filter here operates in a sliding window fashion along the input dimension (details are described below). This layer is followed by a max-pooling operation whose output is then fed into the next convolutional layer.
We extend a single-layer CNN, originally proposed in~\cite{Severyn:2015tb,Kim:2014vt,Kalchbrenner:2014wl}, to two convolutional and pooling layers.
The resulting network architecture is illustrated in Figure \ref{fig:cnn} and in its basic variant consists of two consecutive pairs of convolutional-pooling layers followed by a single hidden layer and a soft-max output layer. In the following, we describe in detail each layer and corresponding parameters.
\vspace{1mm}

\subsection{Convolutional Neural Network}
\label{ssec:cnn}
\vspace{1mm}

\textbf{Embedding layer.}
Each word is associated with a $d$-dimensional vector (embedding). An input sequence of $n$ words is represented by concatenating their embeddings, yielding a sentence matrix $\mathbf{X} \in \rm \R^{d \times n}$.  $\mathbf{X}$ is used as input to the network.
\\

\textbf{Convolutional layer.}
This layer applies a set of $m$ convolutional filters of length $h$ over the matrix $\mathbf{X}$. Let $\mathbf{X}_{[i:i+h]}$ denote the concatenation of word vectors $\mathbf{x}_i$ to $\mathbf{x}_{i + h}$. A feature $c_i$ is generated for a given filter $\mathbf{F}$ by: %
\begin{equation}\label{eq:convolution}
c_i := \sum_{k,j}(\mathbf{X}_{[i:i+h]})_{k,j} \cdot \mathbf{F}_{k,j}
\vspace{-1pt}
\end{equation}
The concatenation of all vectors in a sentence defines a feature vector $\mathbf{c} \in \rm \R^{n-h+1}$. The vectors $\mathbf{c}$ are then aggregated from all $m$ filters into a feature map matrix $\mathbf{C} \in \rm \R^{m \times (n-h+1)}$. The filters are learned during the training phase of the neural network, as described in Section \ref{ssec:pre}.
The output of the convolutional layer is passed through a non-linear activation function, before entering a pooling layer. 
\\

\textbf{Pooling layer.} The pooling layer aggregates the input vectors by taking the maximum over a set of non-overlapping intervals. The resulting pooled feature map matrix has the form: $\mathbf{C_{pooled}} \in \R^{m \times \frac{n-h+1}{s}}$, where $s$ is the length of each interval. In the case of overlapping intervals with a stride value $s_t$, the pooled feature map matrix has the form $\mathbf{C_{pooled}} \in \R^{m \times \frac{n-h+1-s}{s_t}}$. Depending on whether the boundaries are included or not, the result of the fraction is rounded up or down respectively.
\\

\textbf{Hidden layer.}
A fully connected hidden layer computes the transformation $\alpha(\mathbf{W}*\mathbf{x} + \mathbf{b})$, where~$\mathbf{W} \in \rm \R^{m \times m}$ is the weight matrix, $\mathbf{b} \in \rm I\!R^{m}$ the bias, and $\alpha$ the rectified linear (\verb|relu|) activation function~\cite{nair2010rectified}. 
The output $\mathbf{x}\in \rm \R^{m}$ of this layer can be seen as an embedding of the input sentence.
\\

\textbf{Softmax.}
Finally, the outputs of the previous layer $\mathbf{x} \in \rm \R^{m}$ are fully connected to a soft-max regression layer, which returns the class $\hat{y} \in [1,K]$ with largest probability, i.e.,
\begin{align}
\begin{split}
\hat{y} &= \arg \max_j \,P(y=j\ |\ \mathbf{x},\mathbf{w},\mathbf{a}) \\ &= 
\arg \max_j \, \frac{e^{\mathbf{x}^\intercal \mathbf{w}_j + a_j}}{\sum_{k=1}^K e^{\mathbf{x}^\intercal \mathbf{w}_k + a_j}},
\end{split}
\end{align}
where $\mathbf{w}_j$ denotes the weights vector of class~$j$, from which the dot product with the input is formed, and $a_j$ the bias of class $j$. %
\\

\textbf{Network Parameters.}
The following parameters of the neural network are learned during training: $\mathbf{\theta} =$ $\{ \mathbf{X}, \mathbf{F}_1, \mathbf{b}_1,$ $\mathbf{F}_2, \mathbf{b}_2, \mathbf{W}, \mathbf{a} \}$, with $\mathbf{X}$ the word embedding matrix, where each row contains the $d$-dimensional embedding vector for a specific word; $\mathbf{F}_i, \mathbf{b}_i$ the filter weights and biases of convolutional layers; and $\mathbf{W}$ and $\mathbf{a}$ the weight-matrix for output classes in the soft-max layer.
\\

\subsection{Learning the Model Parameters}
\label{ssec:pre}

The model parameters are learned using the following three-phase procedure: (i) creation of word embeddings; (ii) distant-supervised phase, where the network parameters are tuned by training on weakly labelled examples; and (iii) final supervised phase, where the network is trained on the supervised training data.
\\

\textbf{Preprocessing and Word Embeddings.}
The word embeddings are learned on an unsupervised corpus containing \ntweets tweets. We apply a skip-gram model of window-size~5 and filter words that occur less than 15 times~\cite{Severyn:2015tb}. The dimensionality of the vector representation is set to $d=52$. Our experiments showed that using a larger dimension did not yield any significant improvement.
\\

\textbf{Training.}
During the first distant-supervised phase, we use emoticons to infer noisy labels on tweets in the training set~\cite{read2005using,Go:2009ut}. Details of the training set obtained from this procedure are listed in Table~\ref{tbl:data}, "Pre-training" rows. Note that the data used for this pre-training phase do not necessarily overlap with the data used to create the word embeddings. The neural network was trained on these datasets for one epoch, before finally training on the supervised data for about~20 epochs. The word embeddings $\mathbf{X}\in\R^{d\times n}$ are updated during both the distant- and supervised training phases by applying back-propagation through the entire network. The resulting effect on the embeddings of words are illustrated in Figure~\ref{fig:wemb_after} and discussed in Section~\ref{sec:results}.
\\

\textbf{Optimization.}
During both training phases, the network parameters are learned using \emph{AdaDelta}~\cite{Zeiler2012}. We compute the score on the validation set at fixed intervals and select the parameters achieving the highest score.
\\

\begin{figure}
  \center
  \includegraphics[width=0.49\textwidth]{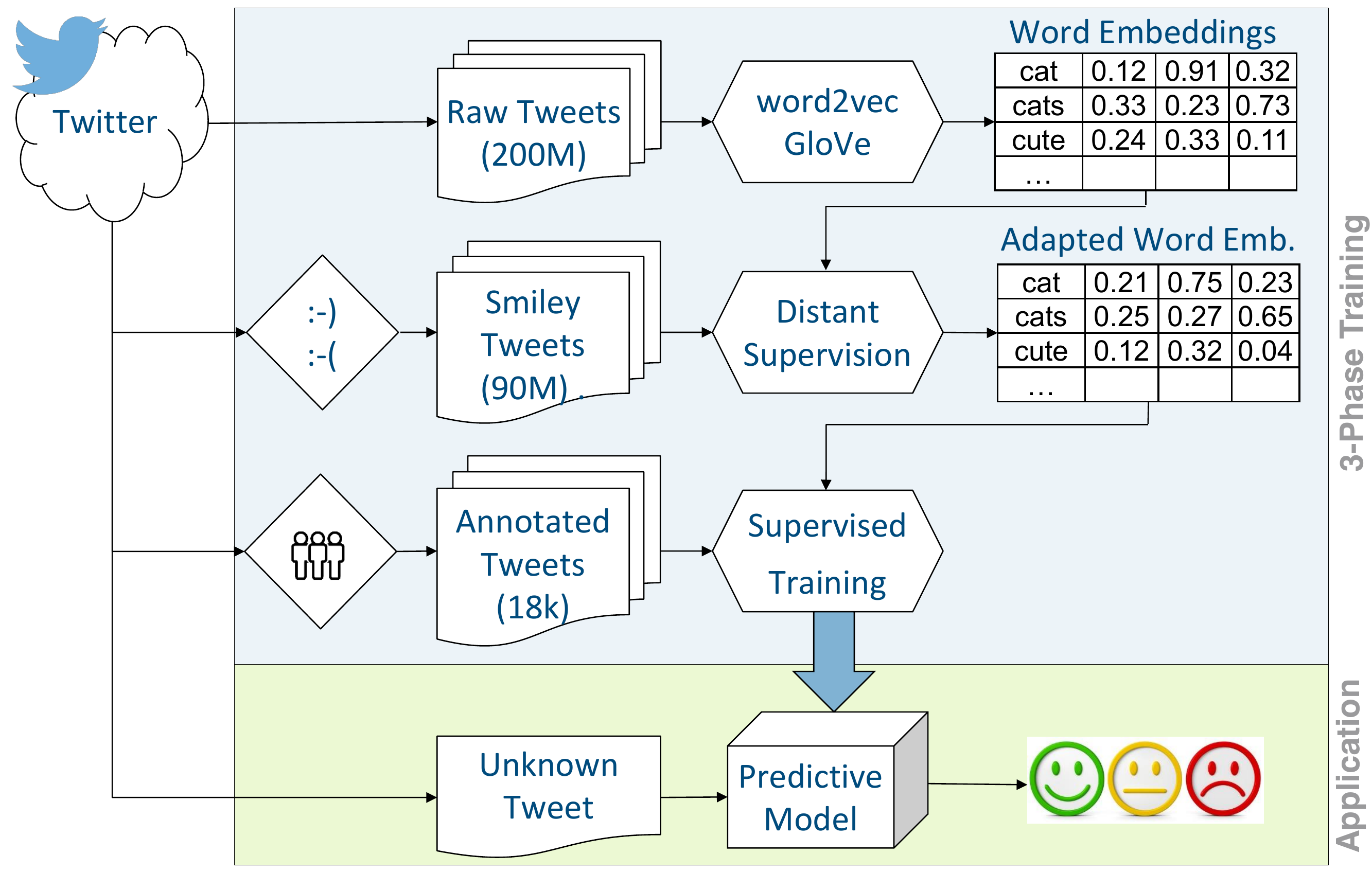}
  \vspace{-1mm}
  \captionof{figure}{\textit{Training Phases Overview.}}
  \vspace{2mm}
  \label{fig:flowChart}
\end{figure}

Figure \ref{fig:flowChart} shows a complete overview of the three phases of the learning procedure.

\section{Experimental Setting}

\begin{table}
\begin{center}\small
\caption{\textit{Data used for training the CNN model.}}
\label{tbl:data}
\begin{tabular}{clrrrr}
\hline
\!\!\textbf{Language}\!\!\!\! & \textbf{Dataset} & \textbf{Total} & \textbf{Neutral} & \textbf{Neg.} & \textbf{Pos.} \\
\hline
\multirow{5}{*}{English} & Word embeddings & 300M & - & - & -\\
& Pre-training & 60M & - & 30M & 30M\\
& Training & 18044 & 7544 & 2748 & 7752 \\
& Validation & 2390 & 987 & 365 & 1038 \\
& Test & 20632 & 10342 & 3231 & 7059 \\
\hline
\multirow{4}{*}{French}& Word embeddings & 300M & - & - & -\\
& Pre-training & 60M & - & 20M & 40M\\
& Training & 9107 & 4095 & 2120 & 2892 \\
& Test & 3238 & 1452 & 768 & 1018 \\
\hline
\multirow{4}{*}{German}& Word embeddings & 300M & - & - & -\\
 & Pre-training & 40M & - & 8M & 32M\\
& Training & 8955 & 5319 & 1443 & 2193 \\
& Test & 994 & 567 & 177 & 250  \\
\hline
\multirow{4}{*}{Italian}& Word embeddings & 300M & - & - & -\\
 & Pre-training & 40M & - & 10M & 30M\\
& Training & 6669 & 2942 & 2293 & 1434 \\
& Test & 741 & 314 & 250 & 177 \\
\hline
\hline
\end{tabular}
\end{center}\vspace{-3mm}
\end{table}

\subsection{Data}
We used a large set of \ntweets tweets to create the word embeddings for each language, as well as a distant supervision corpus of $40$-$60M$ tweets for each language, where each tweet contained at least one emoticon (positive or negative smiley).
Smileys were used to automatically infer weak labels and subsequently removed from the tweets. This idea of distant-supervised learning was described in~\cite{Go:2009ut,Severyn:2015tb,Severyn:2015kt}.
For the final supervised phase, we used publicly available labeled datasets for English~\cite{SemEval:2016:task4}, Italian~\cite{sentipolc2016} and French~\cite{deft2015}. %
The German corpus was newly created by the authors and is available at \url{http://spinningbytes.com/resources}.
An overview of the datasets used in our experiment, including the number of labelled tweets per dataset, is given in Table~\ref{tbl:data}.
\\

\textbf{Data Preparation.}
Each tweet was preprocessed in three steps: (i) URLs and usernames were substituted by a replacement token, (ii) the text was lowercased and (iii) finally tokenized using the NLTK tokenizer.
\\

\subsection{Sentiment Analysis Systems}
\label{sec:baselines}
In our experiments, we compare the performance of the following sentiment analysis systems:
\begin{itemize}
\itemsep0em
\item \textit{Random forest} (RF) as a common baseline classifier. The RF was trained on $n$-gram features, as described in~\cite{narr2012language}
\item \textit{Single-language CNN} (SL-CNN). The CNN with three-phase training, as described in Section~\ref{sec:model}, is trained for each single language. In a set of experiments, the amount of training in the three phases is gradually reduced. The system using all available training data for one language is also referred to as 'fully-trained CNN'
\item \textit{Multi-language CNN} (ML-CNN), where the distant-supervised phase is performed jointly for all languages at once, and the final supervised phase independently for each language. For the pre-training phase, we used a balanced set of \ntweets that included all four languages, see Table \ref{tbl:data}, 'Pre-training' 
\item \textit{Fully multi-language CNN} (FML-CNN), where all training phases were performed without differentiation between languages. The pre-training data is the same as in ML-CNN
\item \textit{SemEval benchmark.} In addition, results on the English dataset were compared to the best known ones from the SemEval benchmark\footnote{\url{http://alt.qcri.org/semeval2016/}}. For the data sets in the other three languages, no public benchmark results could be found in literature
\item \textit{Translate}: this approach uses Google Translate\footnote{\url{https://translate.google.com/}} (as of Oct 2016) to translate each input text from a source language to a target language. It then uses the SL-CNN classifier trained for the target language to classify the tweets %
\end{itemize}

\subsection{Performance Measure}
We evaluate the performance of the proposed models using the metric of SemEval-2016 challenge, which consists in averaging the macro F1-score of the positive and negative classes\footnote{Note that this still takes into account the prediction preformance on the neutral class.}. Each approach was trained for a fixed number of epochs. We then selected the results that yielded the best results on a separate validation set.

For French, German and Italian, we created a validation set by randomly sampling $10\%$ of the data. For English we used the \texttt{test2015} set as validation set and \texttt{test2016} for testing from the SemEval-2016 challenge, see Validation set in Table~\ref{tbl:data}.

\subsection{Implementation Details}
The core routines of our system are written in \verb|Theano|~\cite{Bergstra2010} exploiting GPU acceleration with the \verb|CuDNN| library \cite{Chetlur2014}. The whole learning procedure takes approximately 24-48 hours to create the word embeddings, 20 hours for the distant-supervised phase with $160M$ tweets and only 30 minutes for the supervised phase with $35K$ tweets.

Experiments were conducted on 'g2.2xlarge' instances of \emph{Amazon Web Services} (AWS) with \emph{GRID K520} GPU having 3072 CUDA cores and 8 GB of RAM.
\\

\section{Results}
\label{sec:results}
In this section, we summarize the main results of our experiments. 

The F1-scores of the proposed approach and competing baselines are summarized in Table~\ref{tab:results}. The fully-trained SL-CNNs significantly outperforms the other methods in all four languages. The best F1-score was achieved for Italian (67.79\%), followed by German (65.09\%) and French (64.79\%), while the system for English reached only 62.26\%. The proposed SL-CNNs outperform the corresponding baselines from literature and RF.
\\

\begin{table}[b]
\vspace{2mm}
\caption{\textit{F1-scores of compared methods on the test sets.} The highest scores among the three proposed models are highlighted in bold face. \textbf{ML-CNN} and \textbf{FML-CNN} are two variants of the method presented in Section~\ref{sec:cross-language}.}
\vspace{-2mm}
\label{tab:results}
\begin{center}
\begin{tabular}{l rrrrr}
\toprule
\textbf{Method} 	 & \multicolumn{4}{c}{\textbf{Language}} \\
& \textbf{English} & \textbf{French} & \textbf{German} &  \textbf{Italian}  \\
\hline
\hline
SL-CNN & {\bf 63.49} & {\bf 64.79} &{\bf 65.09} & {\bf 67.79} \\
ML-CNN &  61.61 & - & 63.62 & 64.73 \\
FML-CNN & 61.03 & - & 63.19 & 64.80 \\
\hline
RF & 48.60 & 53.86 & 52.40 & 52.71 \\
SENSEI-LIF~\cite{SemEval:2016:task4}\!\!\!\!\!\!& 62.96 & - & - & - \\
UNIMELB~\cite{SemEval:2016:task4}\!\!\!\!\!\!& 61.67 & - & - & - \\
\hline
\end{tabular}\vspace{-1em}
\end{center}
\end{table}

\begin{figure}
  \center
  \includegraphics[width=0.4\textwidth]{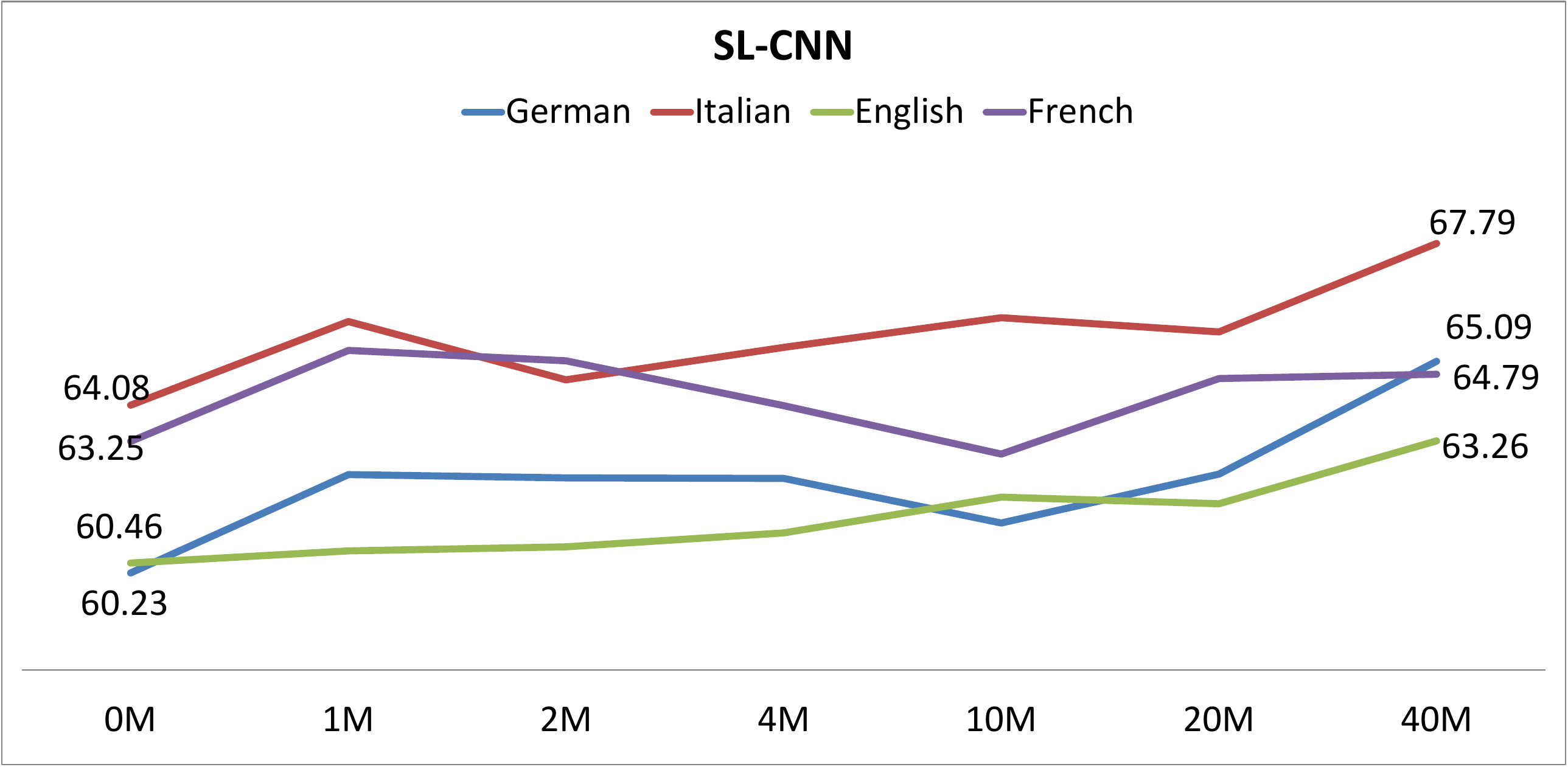}
  \vspace{1mm}
  \captionof{figure}{\textit{Results obtained by varying the amount of data during the distant supervised phase.} Each CNN was trained for one epoch.}
  \label{fig:res1}
\end{figure} 

\textbf{Leveraging Distant Training Data.} 
We increased the amount of data for the distant-supervised phase for SL-CNN. Figure~\ref{fig:res1} compares F1-scores for each language when changing the amount of tweets from 0 to 40M. 
The scores without distant supervision are the lowest for all languages. We observe a general increase of F1-score when increasing the amount of training data. The performance gain for English, Italian and German is around 3\%, while it is more moderate for French.
\\

\begin{figure}
	\begin{center}
		\includegraphics[width=0.4\textwidth]{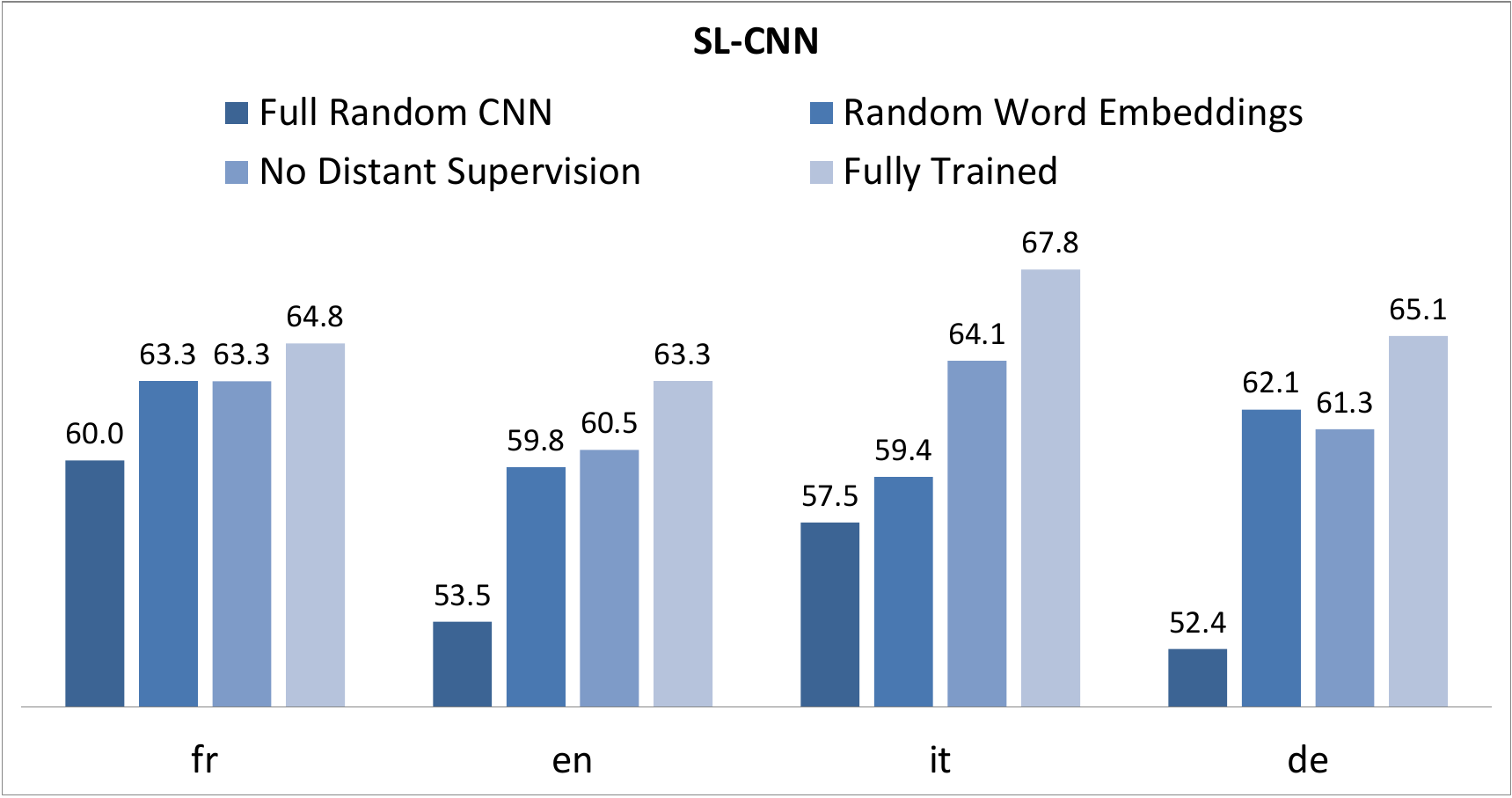}
	\end{center}
	\caption{\textit{Results obtained by initializing the CNNs with different word embeddings.} The fully trained variant typically performs better than the other three variants, thus demonstrating the importance of initializing the words vectors as well as performing distant-supervised training.}
\label{fig:word_embeddings}
\end{figure}

\begin{figure}
  \vspace{2mm}
	\begin{center}
	\includegraphics[width=0.35\textwidth]{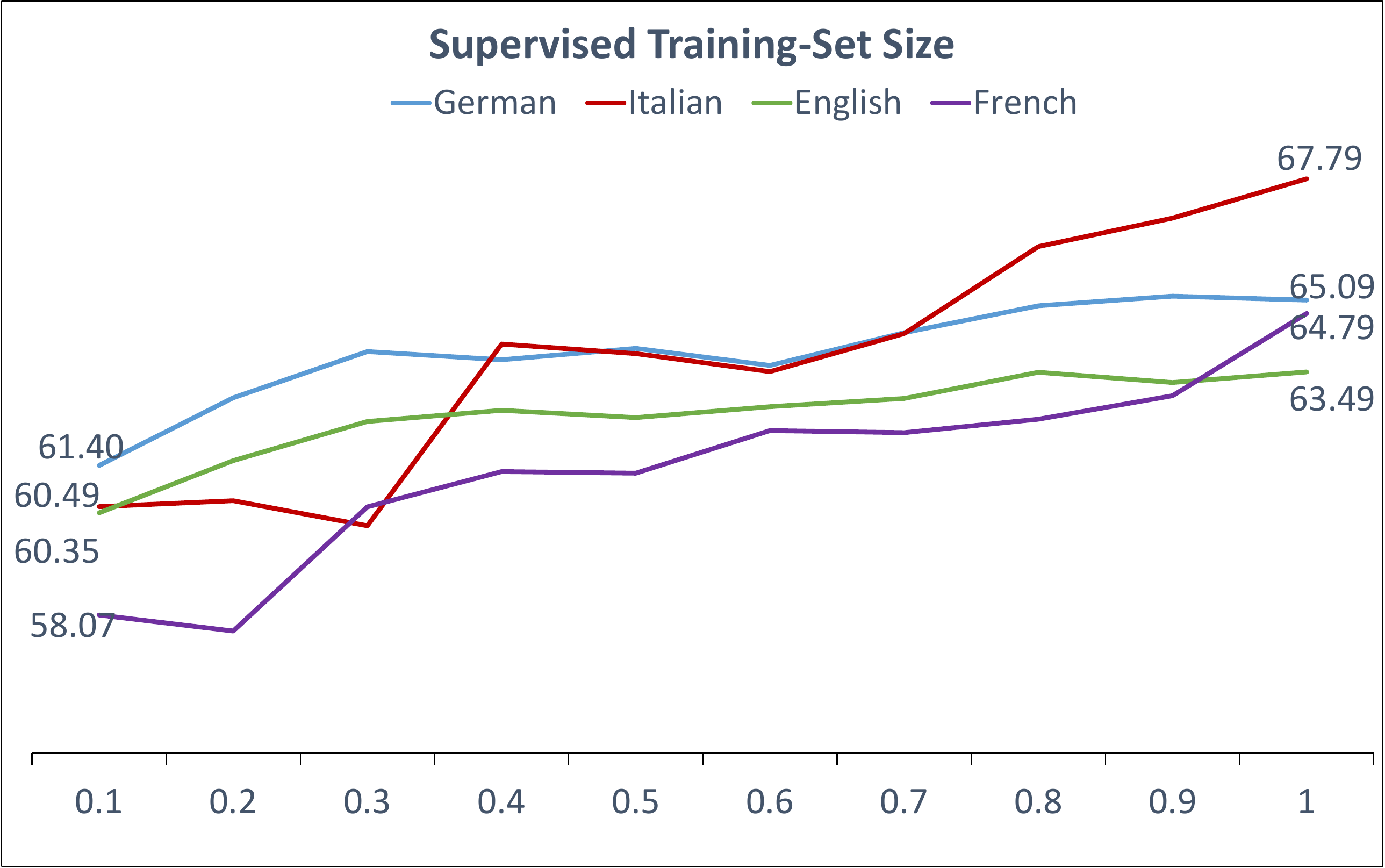}
	\vspace{1mm}
	\caption{\textit{Results obtained by varying the amount of supervised data. The maximum number on the y-axis corresponds to the total number of training tweets given in Table~\ref{tbl:data}.}}
	\label{fig:supervised_data}
	\end{center}
\end{figure}

\textbf{Supervised data.}
In Figure~\ref{fig:supervised_data} we report the F1-scores of each model for increasing amount of supervised data.
We observe a score increase of 2-4\% when using 100\% of the available data instead of 10\%.
\\

\textbf{Word Embeddings.}
We investigate the importance of initialization of word embeddings and the interaction of the latter with the distant supervised phase in four scenarios: 
(i) using randomly initialized word embedding weights, not updated during the distant-supervised phase (named \textit{Full Random CNN}),
(ii) using randomly initialized word embeddings, updated during the distant-supervised phase (\textit{Random Word Embeddings}),
(iii) using word2vec embeddings without distant supervision (\textit{No Distant Supervision}) and
(iv) using word2vec embeddings, updated during the distant-super\-vised phase using $160M$ tweets (\textit{Fully trained CNN}).
Results in Figure~\ref{fig:word_embeddings} demonstrate that the \textit{Fully trained CNN} approach performed the best in almost all cases. These results prove that the quality of initialization as well as updating the large number of word vector parameters during training of the network yield significant improvements.
\\

\begin{table*}
\caption{\captionfont{\textit{Summary of the network parameters used for experimental results}. \textsc{L2} is the default architecture used by our approach. The alternative \textsc{L1} and \textsc{L3} architectures are discussed in Figure~\ref{fig:resLayers}.
}}
\vspace{-1mm}
\begin{center}
\begin{small}
\begin{tabular}{lcccc} 
\hline
\textsc{} & Number of layers & Number of filters & Filter window size $h$ & Size of max-pooling window $w$ and striding $st$ \\
\hline 
\textsc{L1}	 	&$1$&$300$&$h_1 = 5$ & None\\
\textsc{L2}		&$2$&$200$&$h_1 = 4$, $h_2 = 3$ & $w_1=4$, $st_1=2$\\
\textsc{L3}		&$3$&$200$&$h_1 = 4$, $h_2 = 3$, $h_3 = 2$ & $w_1=4$, $st_1=2$, $w_2=3$, $st_2=1$\\
\hline
\end{tabular}
\end{small}
\\~\\
  \vspace{1mm}
\end{center}
\label{table:system_comparison}
\end{table*}

\begin{figure*}
	\begin{center}
        \begin{tabular}{@{}c@{\hspace{5mm}}c@{\hspace{5mm}}c@{}}
		\includegraphics[width=0.32\textwidth]{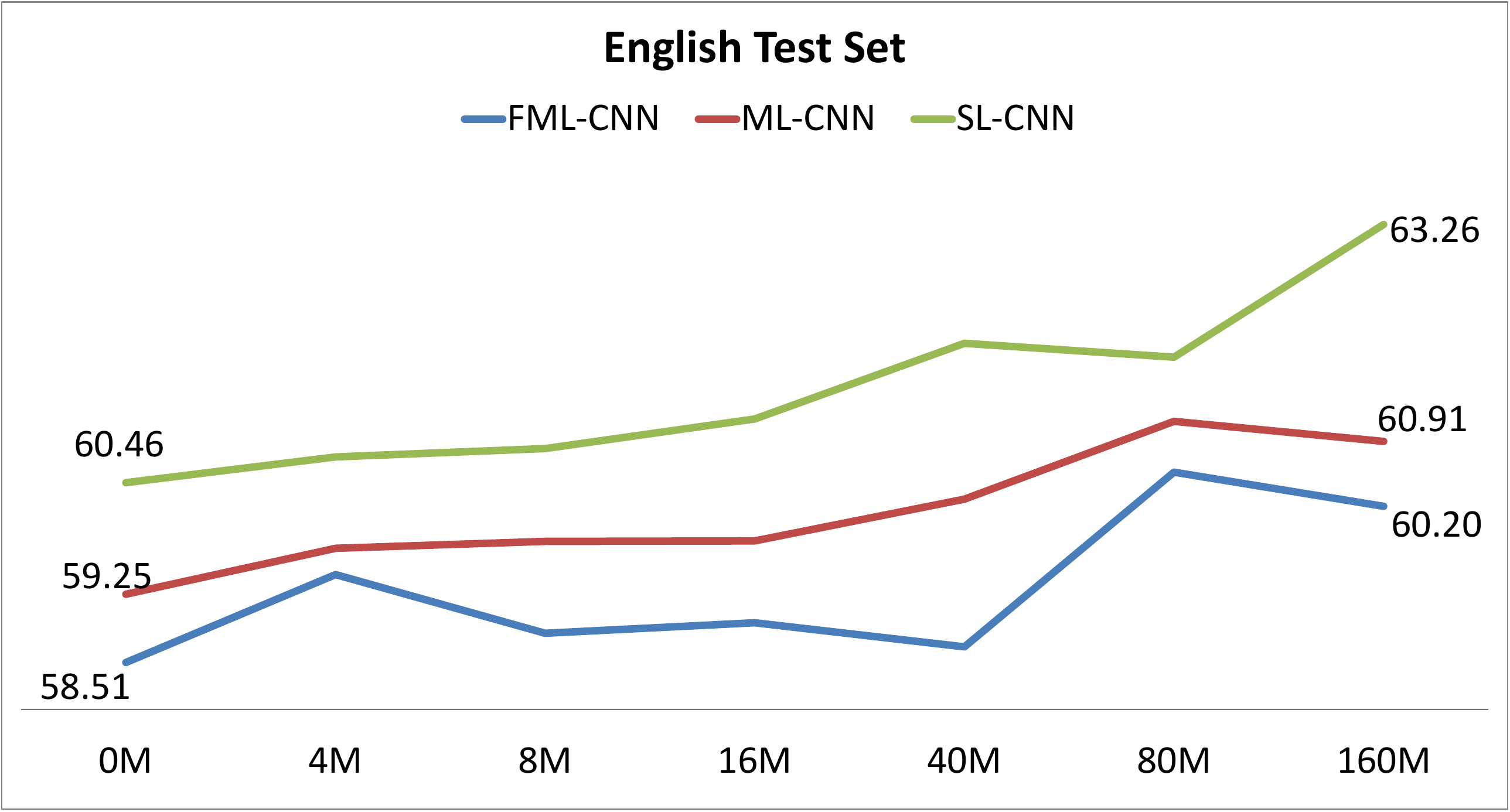} &
		\includegraphics[width=0.32\textwidth]{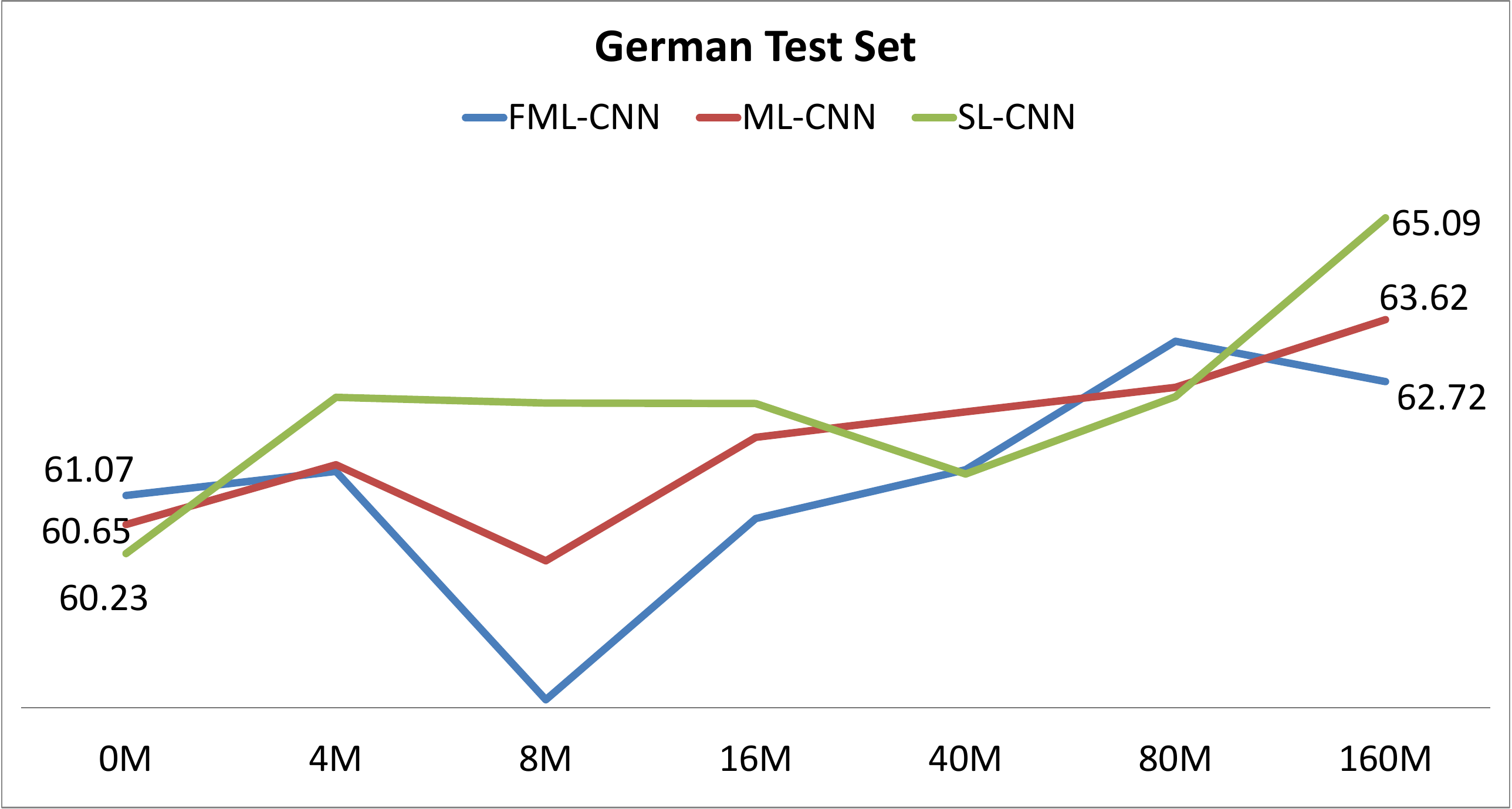} &
		\includegraphics[width=0.32\textwidth]{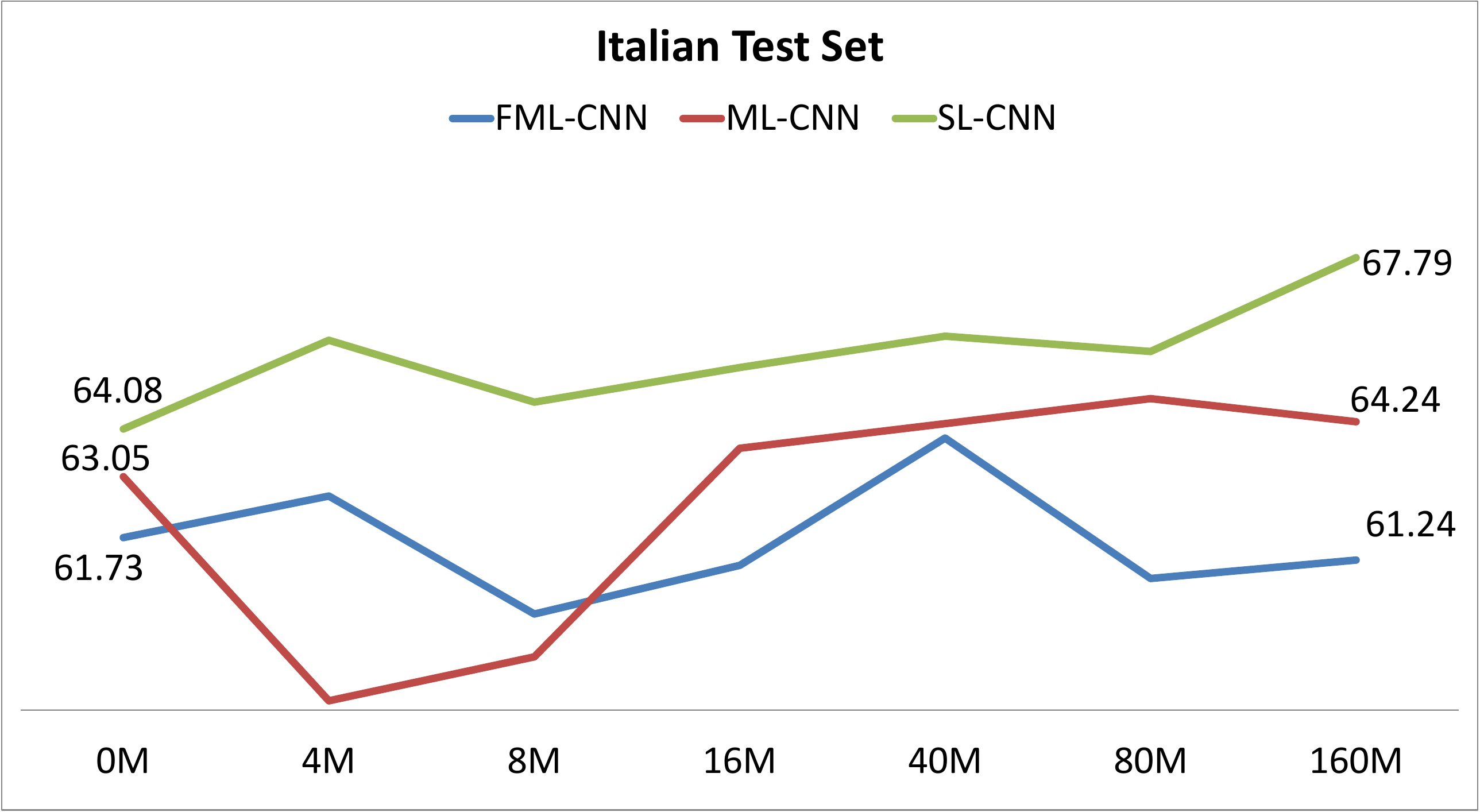} \\	%
	    \end{tabular}
	\end{center}
	\caption{\textit{Results obtained by varying the amount of data during the distant supervised phase.} Each CNN was trained for one epoch. We rescaled the curve for~\textit{SL-CNN} to match the amount of data used per language by the multi-language approaches. For example, while the multi-language approaches were trained with $40M$ tweets ($10M$ tweets per language), each \textit{SL-CNN} model was trained with $10M$ tweets from the same language. Each experiment set, up to $160M$ tweets, contains the same number of tweets per language.
	\vspace{2mm}}
\label{fig:res_ml}
\end{figure*}

\begin{figure}
	\begin{center}
        \begin{tabular}{@{}c@{}}
		\includegraphics[width=0.4\textwidth]{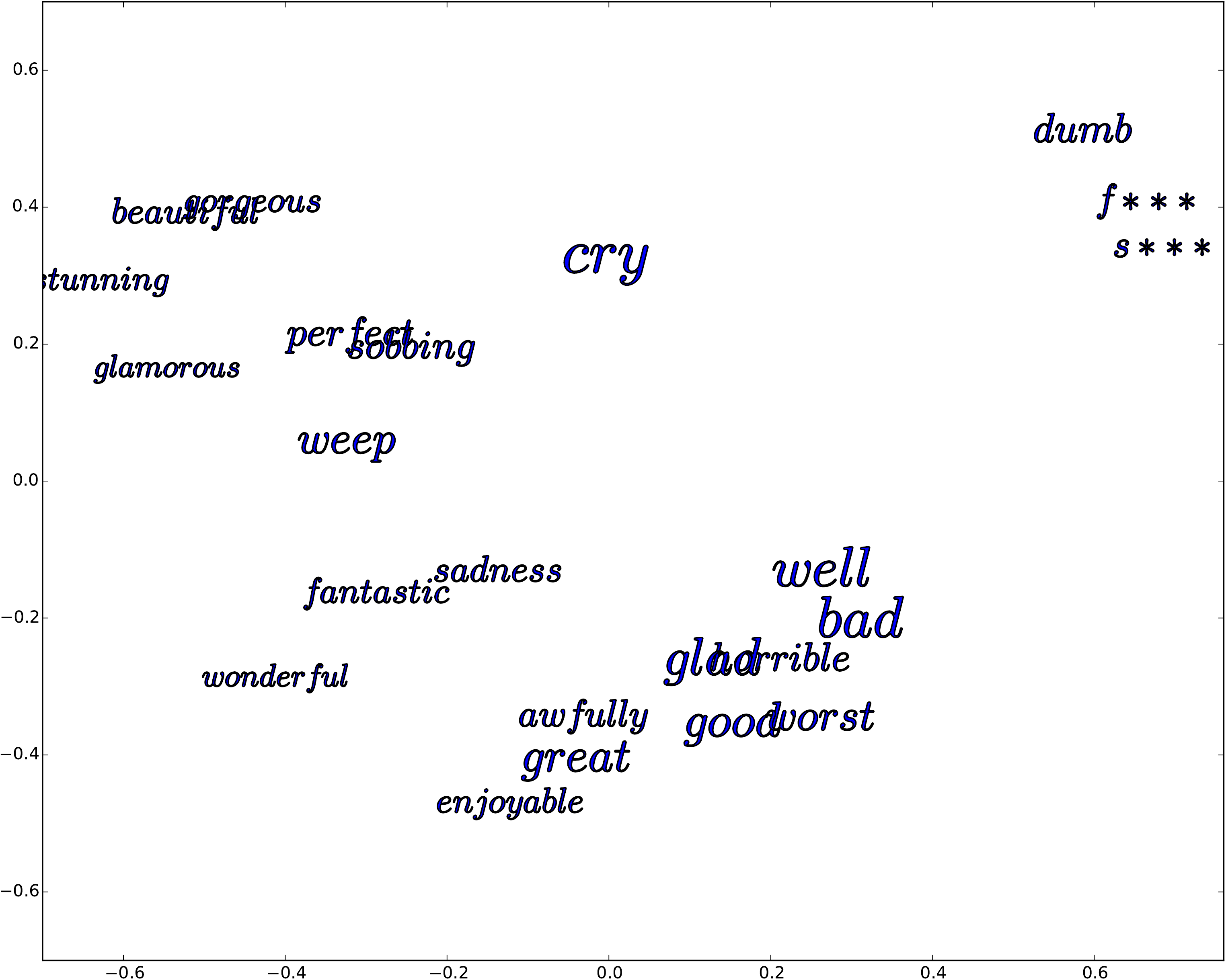} \\
		(a) \vspace{2mm}\\
		\includegraphics[width=0.4\textwidth]{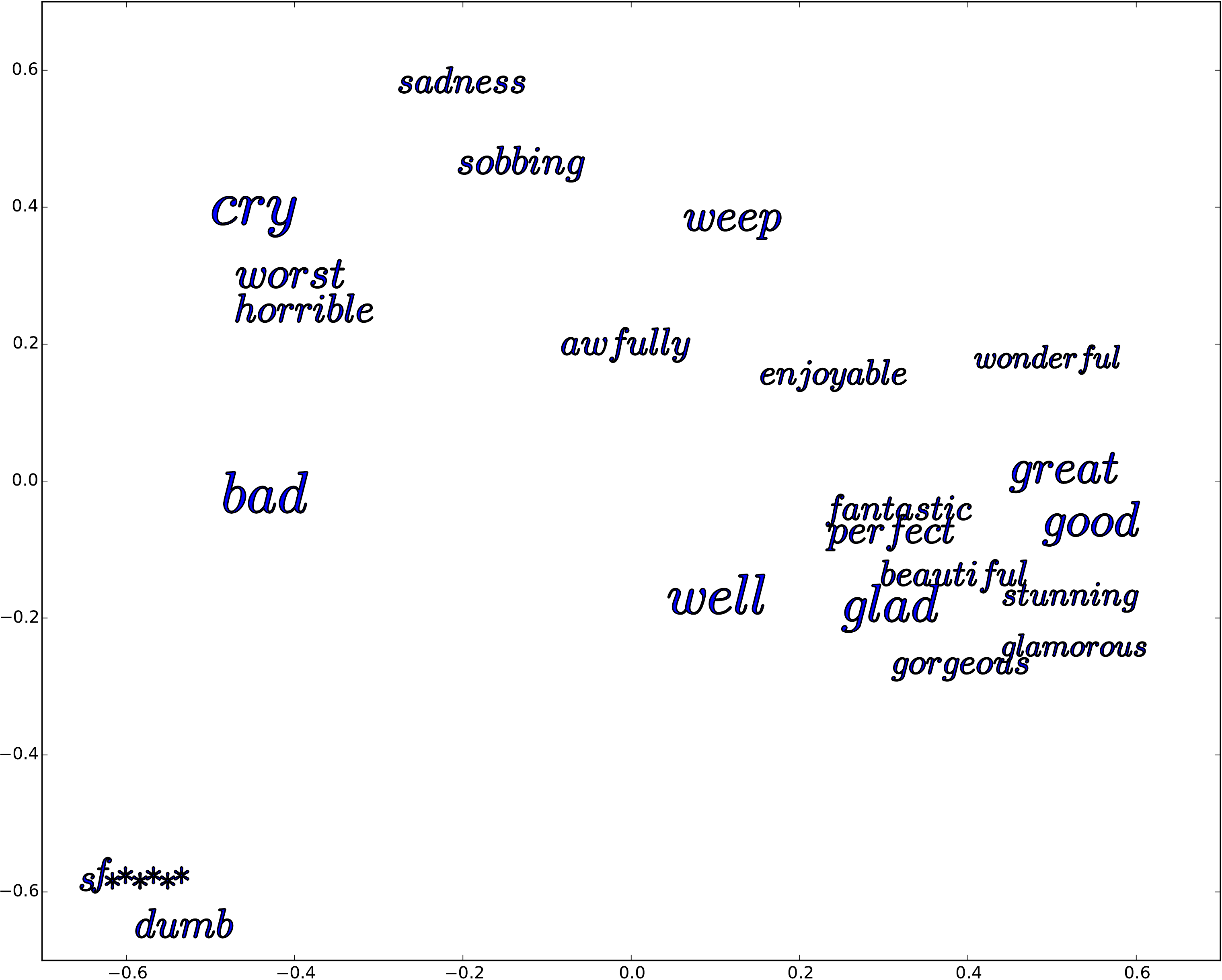} \\
		(b) \\
       \end{tabular}
	\end{center}\vspace{-3mm}
	\captionof{figure}{\textit{Word embeddings projected onto a 2-dimensional space using PCA before (a) and after (b) the distant-supervised training phase.}}
\label{fig:wemb_after}
\end{figure}

\begin{figure}
  \center
  \includegraphics[width=0.48\textwidth]{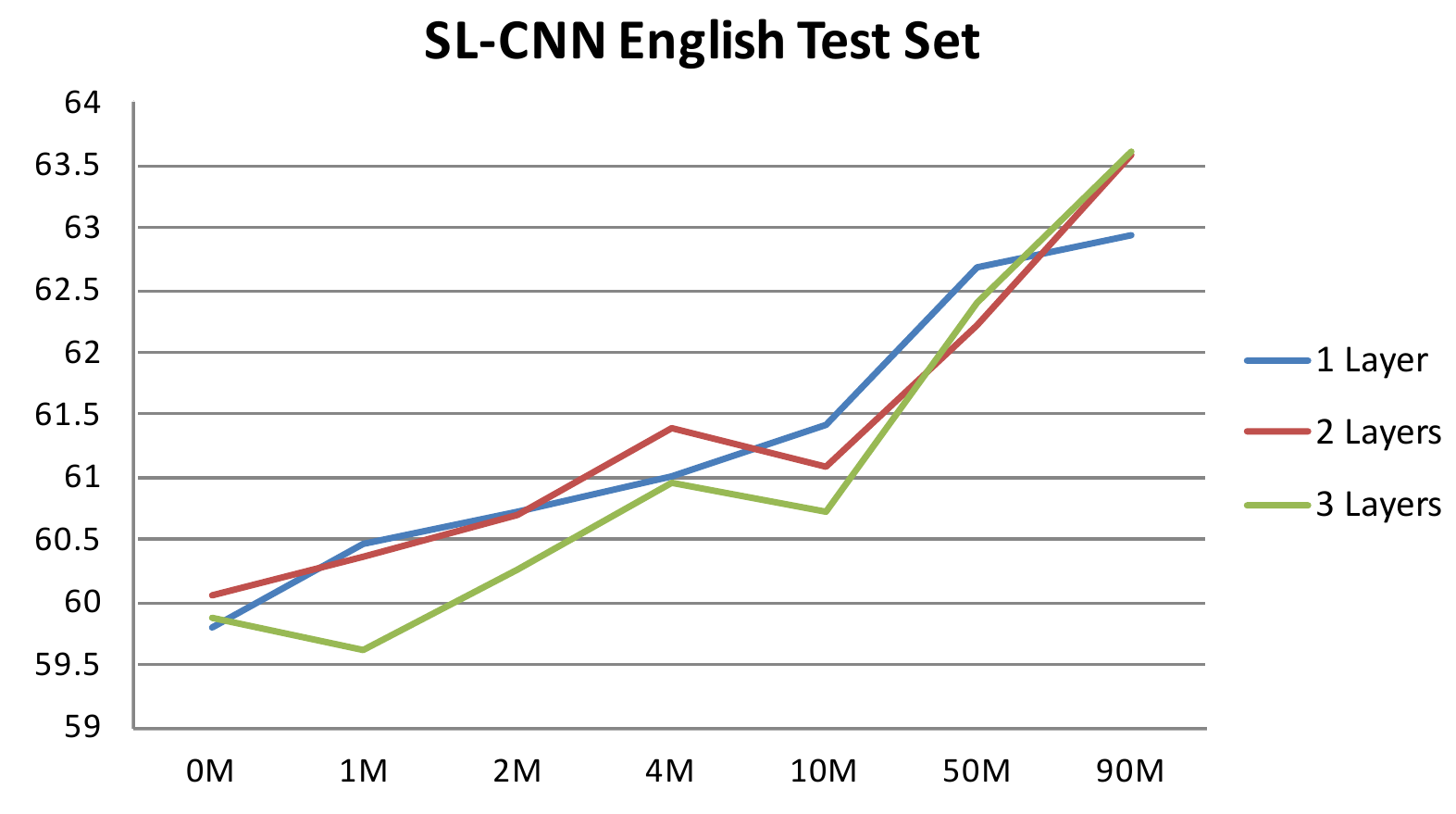}
  \captionof{figure}{\textit{F1-score on the English corpus as a function of the number of network layers.} These results were obtained by training the \textit{SL-CNN} models with different number of layers, as specified in Table~\ref{table:system_comparison}, on different amounts of data during the distant-supervised phase. Each CNN was trained for one distant epoch.
  \vspace{2mm}
  }
  \label{fig:resLayers}
\end{figure}

Figure~\ref{fig:wemb_after} illustrates the effect of the distant-supervised phase on the word embeddings. For visualization purposes, principal component analysis (PCA) was used to project the word embeddings onto two dimensions. We see that the geometry of the word embeddings reflects the distance in terms of sentiment between pairs of words. Figure~\ref{fig:wemb_after}(a) shows the initial word embeddings created by word2vec, before the distant-supervised phase. Taking as an example the pair of words "good" and "bad", it is clear that these two words often appear in the same context and are thus close to each other in the embedded space. The similarity score of these two vectors is $0.785$. After the distant-supervised phase, the semantic of the space is changed and the distance between words come to reflect the difference in terms of sentiment. As shown in Figure~\ref{fig:wemb_after}(b), negative and positive words are neatly separated into two clusters. In this case, the similarity score between the word embeddings of "good" and "bad" becomes $-0.055$. Finer grained clusters are also revealed in the second embedding. For example, words that convey sadness are close together.
\\

\textbf{Comparing Network Architectures.} 
One common question asked by practitioners relates to the influence of the number of layers on the performance of a neural network. We thus evaluated the performance of various architectures with multiple layers. In order to reduce the number of experiments, we evaluated the performance of the single-language model SL-CNN on the English set and varied the number of convolutional/pooling layer pairs from 1 to 3. We evaluated a total of 12 networks. Here, we only report the best set of parameters for each number of layers in Table~\ref{table:system_comparison} and corresponding F1-scores in Figure~\ref{fig:resLayers}. The network performance generally improves with the number of layers, if a sufficient amount of training data is used in the distant-supervised phase. For the task of sentiment classification, current recurrent architectures such at LSTMs still do not perform as well as CNNs, see e.g. \cite[Table 4]{Semeniuta:2016ww} and the discussion in the related work section.
\\

\begin{table}
\caption{\textit{Translation experiment.} We translated each source language to each target language and used the model trained on the target language to predict tweets polarity.}
\vspace{-4mm}
\label{tbl:translation_performance}
\begin{center}
\begin{tabular}{l rrrr}
\toprule
\textbf{Source} 	 & \multicolumn{4}{c}{\textbf{Target}} \\
& English &  Italian & German &  French  \\
\hline
\hline
English & \textbf{63.49} & 59.87 & 57.53 & 58.47 \\
Italian & 64.37	& \textbf{67.79} & 61.57	& 60.19 \\
German & 61.86 & 61.66 & \textbf{65.09} & 61.22 \\
French & \textbf{65.68} & 63.23 & 61.5 & 64.79 \\
\hline
\end{tabular}\vspace{-1mm}
\end{center}
\end{table}

\textbf{Translation Approach.}
In Table~\ref{tbl:translation_performance} we report results of the translation experiment described in Section~\ref{sec:baselines}. The F1-score is higher when not translating tweets to another language for English, Italian and German. As an exception, we obtained better results when translating French to English and using the English model to predict sentiments. %
\\

\subsection{Comparison to multi-language classifiers}
\label{sec:cross-language}

\begin{figure}
	\begin{center}
		\includegraphics[width=0.45\textwidth]{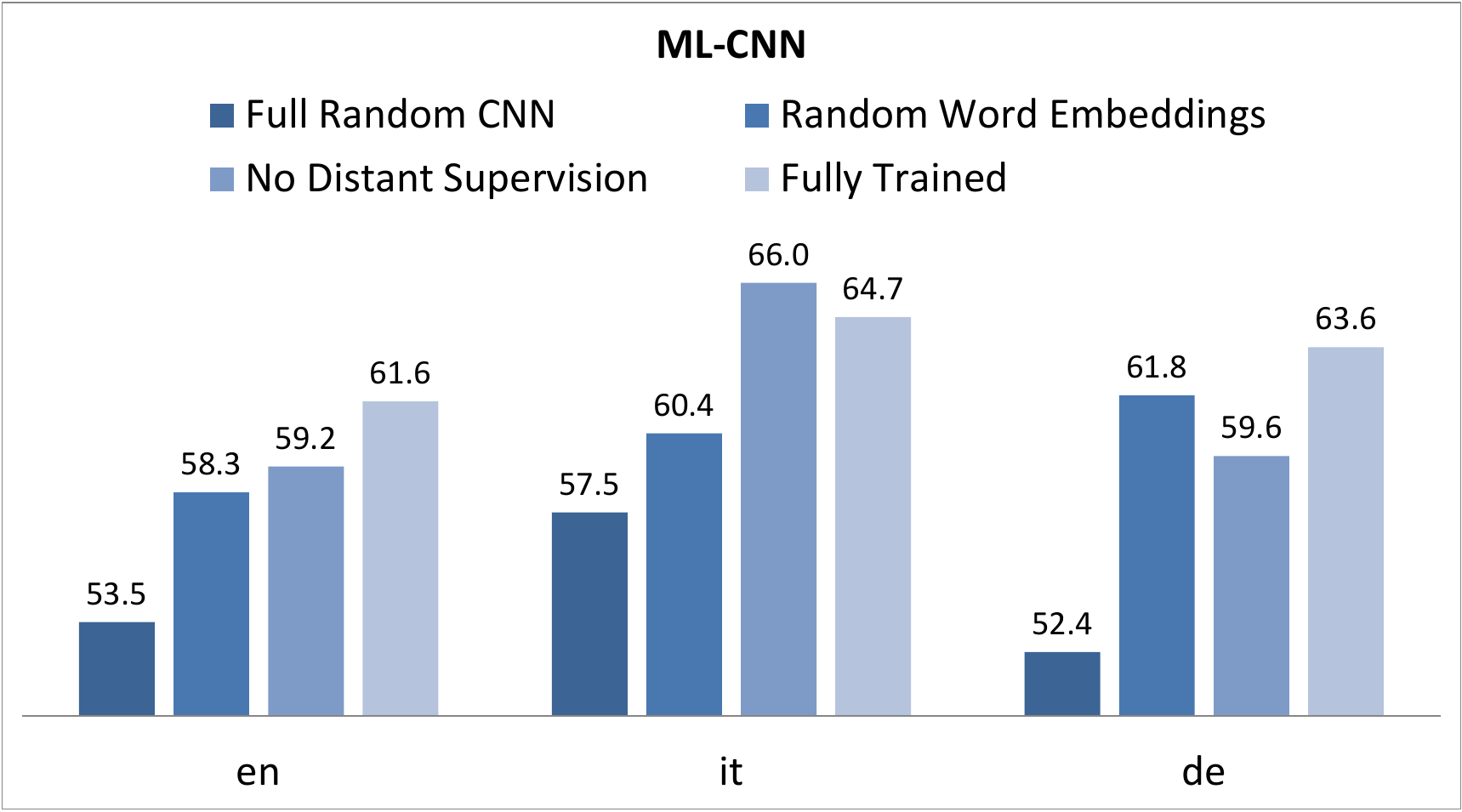} \\
		\includegraphics[width=0.45\textwidth]{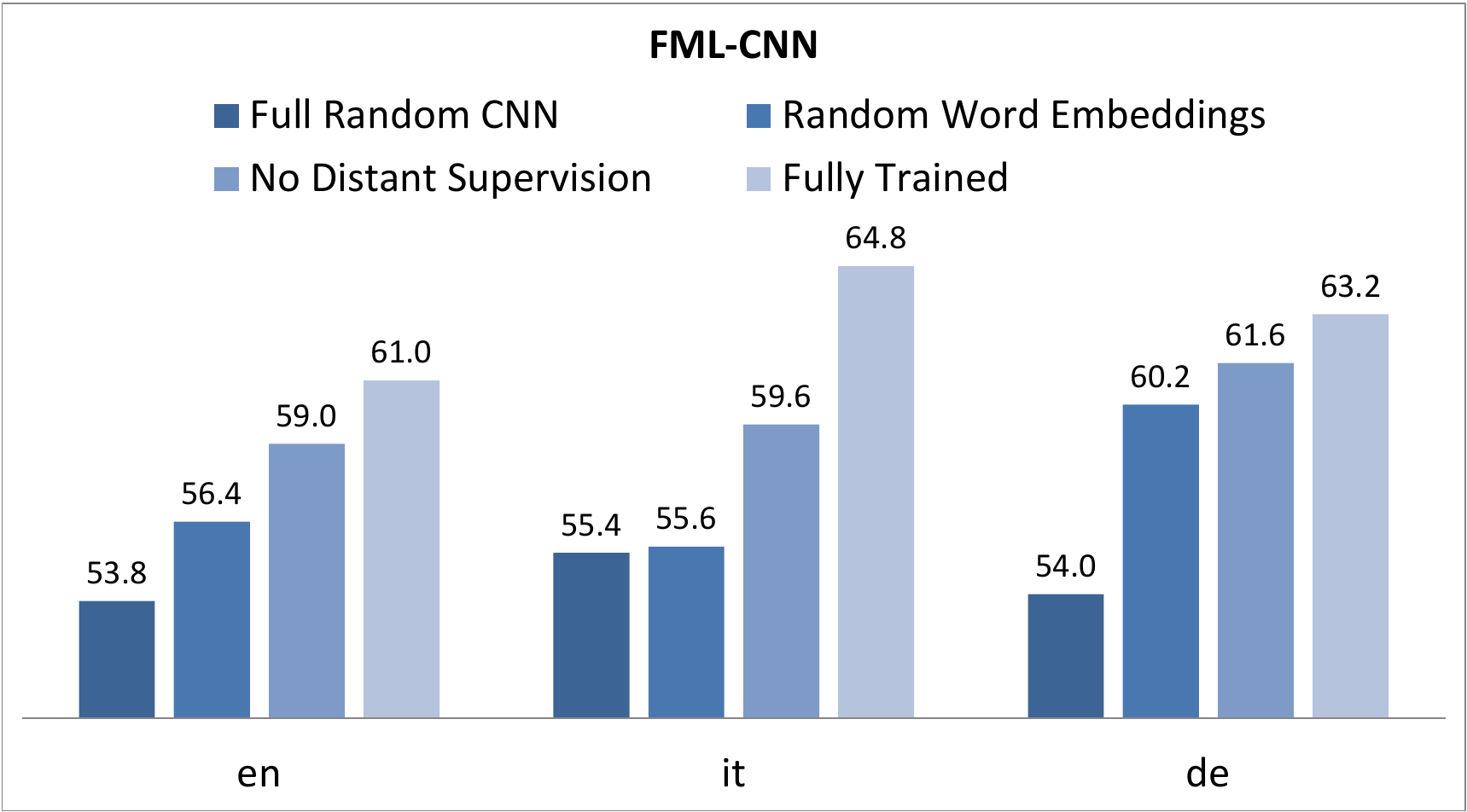} \\
	\end{center}
	\caption{\textit{Results obtained by initializing the CNNs with different word embeddings.} As for the results obtained with the \textit{SL-CNN} model, the fully trained variant typically performs better than the other three variants, both for the \textit{ML-CNN} and \textit{FML-CNN} models.
\\
~
\\
}
\label{fig:word_embeddingsML}
\end{figure}

Figure \ref{fig:res_ml} summarizes F1-scores of the three CNN variants described in Section \ref{sec:baselines}, namely \textit{SL-}, \textit{ML-} and \textit{FML-CNN}, when varying the amount of distant-supervised phase. 
When comparing the three CNN variants, we see that \textit{SL-CNN} gets slightly better scores than \textit{ML-CNN} and \textit{FML-CNN}. The difference in performance between the single and multi-language models is around 2\% on average. However, one benefit of the multi-language models over the single-language ones is their ability to deal with text in mixed languages. To check this hypothesis, we used the~\textit{langpi} tool~\cite{lui2012langid} to extract a set of 300 tweets from the German corpus containing English words. Although these tweets were classified by Twitter as German, they contain a significant number of English words (some of them entirely written in English). We also manually inspected this set and discarded tweets that did not contain English. We then retrained the two models on the training set from which we first removed the set of 300 tweets. When evaluating on this subset, \textit{ML-CNN} obtained an F1-score of 68.26 while \textit{SL-CNN} obtained 64.07. When manually inspecting the results, we clearly observed that \textit{ML-CNN} was better at classifying tweets that were entirely in English or contained several English words. 
The effect of using different word embedding initializations in the multilingual networks is summarized in Figure~\ref{fig:word_embeddingsML}.

\section{Conclusion}

We described a deep learning framework to predict the sentiment polarity of short texts written in multiple languages. In contrast to most existing methods, our approach does not rely on establishing a correspondence to English but instead exploits large amounts of weakly supervised data to train a multi-layer CNN directly in the target language. Through a thorough experimental evaluation, we addressed some fundamental questions around the performance of such model. First, we demonstrated that the strategy used to train these models plays an important role in the obtained performance. Two important factors are a good initialization for the word vectors as well as pre-training using large amounts of weakly supervised data. Second, we compared the performance of a single-language and a multi-language approach. The single-language model reaches the best performance and it even outperforms existing state-of-the-art methods on all the datasets of the SemEval-2016 competition. The multi-language approach performs comparably well or slightly worse than its single-language counterpart, while exhibiting several advantages: %
it does not need to know \textit{a priori} the language(s) used in each tweet; the model can be easily extended to more languages; and it can cope with texts written in multiple languages.
\\

\paragraph{Acknowledgments} 
This research has been funded by Commission for Technology and Innovation (CTI) project no. 18832.1 PFES-ES, and by SpinningBytes AG, Switzerland.
\\
~
\\

\bibliography{bibliography}
\bibliographystyle{abbrv}

\end{document}